\documentclass[lettersize,journal,pagebackref,breaklinks,colorlinks,citecolor=blue]{IEEEtran}
\usepackage{amsmath,amsfonts}
\usepackage{algorithmic}
\usepackage{algorithm}
\usepackage{array}
\usepackage[caption=false,font=normalsize,labelfont=sf,textfont=sf]{subfig}
\usepackage{textcomp}
\usepackage{stfloats}
\usepackage{url}
\usepackage{verbatim}
\usepackage{graphicx}
\usepackage{comment}
\usepackage{cite}
\usepackage{booktabs}
\usepackage{bm} 
\usepackage{color}
\usepackage{pifont}
\usepackage{multirow}
\usepackage{relsize}
\usepackage{caption}
\usepackage[accsupp]{axessibility}
\usepackage{orcidlink}
\begin{document}

\title{Towards Voxel Spacing Consistency for Medical Image Segmentation}

\author{Xin You, Runze Yang, Minghui Zhang, Hanxiao Zhang, Han Li, Yi Yu, Jie Yang,~\IEEEmembership{Senior Member, IEEE}, \\ Nassir Navab,~\IEEEmembership{Fellow, IEEE}, and Yun Gu,~\IEEEmembership{Member, IEEE}
\thanks{This work was supported in part by National Key R\&D Program of China (2022ZD0212400), Shanghai Municipal Commission of Economy and Informatization (2025-GZL-RGZN-BTBX-02023) and Suzhou Industrial
Park Oriental Huaxia Cardiovascular Health Research Institute (GUSU2023002). (Corresponding authors: Yun Gu; Yi Yu.)}
\thanks{X. You, R. Yang, M. Zhang, H. Zhang, J. Yang and Y. Gu are with the Institute of Medical Robotics, Shanghai Jiao Tong University,
200240 Shanghai, China (Email: \{sjtu\_youxin, runze.y, minghuizhang, hanxiao.zhang, jieyang\}@sjtu.edu.cn, yungu@ieee.org).}
\thanks{Y. Yu is with the Department of Ultrasound, Shanghai Chest Hospital, 200052 Shanghai, China (Email: yuyiyy2021@163.com).} 
\thanks{H. Li and N. Navab are with CAMPAR, Technical University of Munich,
80333 Munich, Germany (Email: \{tum\_han.li, nassir.navab\}@tum.de).}
}

\markboth{Journal of \LaTeX\ Class Files,~Vol.~14, No.~8, August~2026}%
{Shell \MakeLowercase{\textit{et al.}}: A Sample Article Using IEEEtran.cls for IEEE Journals}


\maketitle

\begin{abstract}
Volumetric medical image segmentation is essential for both preoperative diagnosis and intraoperative guidance. While recent years have witnessed rapid progress in segmentation architectures, comparatively little attention is paid to the physical voxel spacing of anatomical data. Indeed, volumetric image resampling is a ubiquitous preprocessing step before segmentation, yet its interaction with downstream segmentation has not been systematically exploited. In this work, we study the correlation between image resampling and segmentation, and propose \textbf{Consispace}, a semantic-aware resampling framework that achieves consistent voxel spacing in the axial direction while preserving anatomical and semantic consistency. Consispace introduces an ODE-based anatomical constraint to model inter-slice dynamics with a continuous interpolator, enabling faithful reconstruction under complex anatomical transitions beyond discrete interpolation. To further couple resampling with segmentation objectives, we leverage dense features from a pretrained vision model to build intra-slice semantic correlation maps and inject class-wise semantic consistency via feature reweighting during resampling. Both intra-slice and inter-slice constraints are integrated into an implicit neural network, supporting arbitrary-scale resampling. Extensive experiments on multiple datasets demonstrate that Consispace achieves superior reconstruction quality and perceptual fidelity, produces smoother inter-slice anatomy, and improves downstream segmentation performance when used as a preprocessing step. Codes are available at \url{https://github.com/AlexYouXin/Consispace}.
\end{abstract}



\section{Introduction}
Volumetric medical image segmentation plays a significant role in numerous clinical applications \cite{isensee2021nnu}, including pre-operative diagnostic assistance \cite{bernard2018deep}, treatment planning \cite{nikolov2018deep}, intra-operative guidance \cite{hollon2020near}, and longitudinal assessment of tumor progression \cite{kickingereder2019automated}. To achieve precise and robust segmentation for anatomical structures, existing approaches predominantly emphasize increasingly sophisticated network architectures \cite{tang2022self, isensee2021nnu, roy2023mednext, you2024learning} and anatomically guided loss functions \cite{kervadec2021boundary, sun2023boundary, you2026towards, ma2021loss}. In addition, recent foundation models, such as MedSAM2 \cite{ma2025medsam2} and SAM-Med3D \cite{wang2025sam}, have sought to investigate domain transfer from natural images to the medical imaging domain. However, comparatively little attention has been paid to intrinsic imaging properties, such as the physical voxel spacing of anatomical data, especially in the axial direction.

A recent study, Hyperspace \cite{joutard2024hyperspace}, presents a quantitative analysis of how axial voxel spacing influences the performance of medical image segmentation models. Specifically, the most straightforward solution is to deal with images at their native resolution. However, Hyperspace claims that unifying the slice spacing for training and testing datasets can alleviate the inconsistency of physical spacing, thereby enhancing models' consistent representation of anatomical structures. Notably, this finding is in line with the design choice adopted by state-of-the-art frameworks such as nnU-Net \cite{isensee2021nnu} and TotalSegmentator \cite{wasserthal2023totalsegmentator}. Specifically, nnU-Net resamples data linearly to the median voxel spacing, to realize spacing consistency across all samples. Following this practice, almost all approaches adopt the standard linear interpolation strategy to achieve data harmonization during the preprocessing stage. 

\begin{figure}[!t]
\centerline{\includegraphics[width=1.02\linewidth]{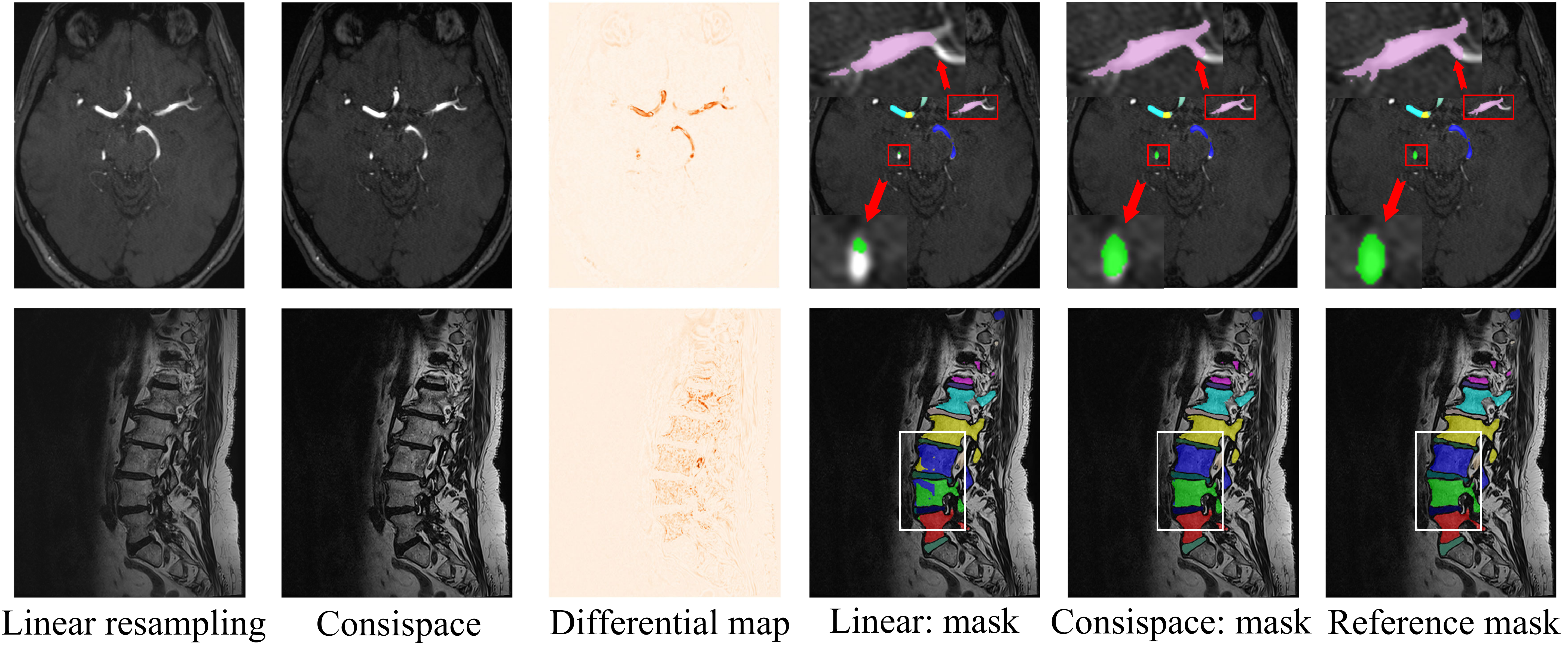}}
  \caption{The performance comparison between linear interpolation and the proposed Consispace on medical image resampling and segmentation. Differential map corresponds to the subtraction of resampled images by linear interpolation and Consispace.
  }
  \label{comparison}
\end{figure}

However, this naturally raises an important question: ``\textbf{Is the current practice of linear resampling truly the optimal strategy?}'' Medical image resampling couples the voxel spacing in the real world with discrete image resolutions, but it fundamentally performs voxel-wise downsampling and upsampling on volumetric data \cite{isensee2021nnu}. Downsampling is less problematic, While upsampling constitutes an ill-posed inverse problem—inferring missing information from limited observations. Consequently, linear interpolation cannot perfectly model complicated inter-slice shape and texture variations, exhibiting two limitations. Firstly, linear interpolation may cause anatomical non-smoothness along the axial direction, leading to structural discontinuities (Fig.\ref{comparison}: the $1^{st}$ row). Secondly, it can result in distorted boundaries between different structures, violating semantic consistency across class-wise anatomies (Fig.\ref{comparison}: the $2^{nd}$ row) \cite{you2025slord}. Furthermore, upsampling in this work focuses on interpolation along the axial direction. Essentially, it is analogous to volumetric super-resolution (VSR) \cite{peng2020saint}, but differs from conventional 2D super-resolution tasks \cite{chen2021learning}. However, prior VSR models \cite{peng2020saint, wu2022arbitrary, yu2022rplhr, fang2024cycleinr, lyu2025diffusion} treat resampling as an upstream task independent of downstream segmentation, thereby ignoring the intrinsic task correlation.

In this paper, we pioneer the exploration of the correlation between volumetric image resampling and medical image segmentation. On the one hand, to boost the segmentation performance on volumetric images, an effective image resampling framework is designed to unify the physical spacing distribution.  Specifically, we propose an ordinary differential equation (ODE)-based anatomical constraint in the resampling network, to characterize the inter-slice spatial dynamics through an ODE-based interpolator. This interpolator enables the network to describe complex anatomical transitions in a continuous and differentiable manner, thereby mitigating the limitations of conventional discrete modeling approaches. Meanwhile, it will boost segmentation through the unified voxel spacing, as demonstrated by the aforementioned study \cite{isensee2021nnu, joutard2024hyperspace}.

On the other hand, segmentation guidance is effective in promoting the process of medical image resampling. Concretely, we introduce semantic consistency into class-wise anatomies. On the ground that pretrained vision models provide high-quality dense features, semantic correlation maps can be extracted to encode contextual relations within each slice. Further, the boundaries of class-wise anatomies are clearly delineated semantically, which will enhance segmentation performance in the downstream task. Both intra-slice and inter-slice constraints are embedded into an implicit neural representation (INR) framework \cite{chen2021learning, wu2022arbitrary} to achieve image resampling at random scales.

The proposed model naturally generalizes on multiple datasets by unifying data with heterogeneous voxel spacings into a Consistent voxel space. Thus, we nominate this resampling framework as \textbf{Consispace}. Experiments demonstrate that Consispace achieves promising image reconstruction performance and feature-level visual perceptual effects. Qualitative results illustrate that Consispace reveals resampled volumetric images with better inter-slice anatomical smoothness. Besides, in the downstream segmentation evaluation based on nn-UNet \cite{isensee2021nnu}, Swin UNETR \cite{tang2022self} and recent MedSAM-2 \cite{zhu2024medical}, datasets preprocessed by Consispace bring more significant increases compared to standard linear resampling. Our main contributions are summarized as follows:
\begin{itemize}
\item We present the first systematic investigation of the bi-directional interaction between volumetric data resampling and medical image segmentation.
\item Firstly, an INR-based resampling pipeline, Consispace, is introduced to harmonize the physical spacing distribution, which will strengthen segmentation performance in the downstream task. To achieve uniform physical spacing, we introduce an ODE-based anatomical constraint in the resampling network to boost inter-slice smoothness through a continuous formulation. 
\item Secondly, class-wise semantic consistency is enforced to improve medical image resampling. Intra-slice semantic correlation maps are derived from a pretrained vision model to produce semantically coherent anatomical structures.
\item Extensive results show that Consispace achieves superior resampling quality, yields inter-slice anatomical continuity, and consistently improves downstream segmentation performance when used as a preprocessing step.
\end{itemize}

\section{Related Work}
\subsection{Medical Image Segmentation}
U-shaped encoder–decoder architectures \cite{ronneberger2015u} have dominated this field. Firstly, convolutional neural networks (CNNs) \cite{cciccek20163d, isensee2021nnu, roy2023mednext, he2016deep} are fully leveraged due to the inductive bias and locality properties of convolutional kernels. In particular, 3D U-Net variants and strong engineering-oriented baselines such as nnU-Net \cite{isensee2021nnu} have demonstrated robust performance across diverse anatomical structures, largely due to their effective multi-scale feature aggregation and carefully designed training pipelines. Further, Transformer-based models \cite{valanarasu2022unext, chen2021transunet, chen2024transunet, tang2022self, zhou2023nnformer} have been introduced to enhance global context modeling in 3D volumes, typically by combining self-attention encoders \cite{dosovitskiy2020image} with U-shaped decoders. Concretely, Swin UNETR \cite{tang2022self} introduced the windowed attention mechanism to alleviate the cubic complexity in volumetric settings. Since self-attention incurs quadratic computational costs with respect to the token number, Mamba-based networks \cite{cheng2025mamba, zhang2025switch, xing2025segmamba, liu2024swin, ma2024u} have been proposed to model the global context within medical images with linear-time complexity. More recently, foundation-model paradigms represented by Segment Anything Model (SAM) \cite{kirillov2023segment} have shifted attention toward promptable segmentation and cross-domain generalization; medical adaptations \cite{ma2025medsam2, wu2025medical, zhu2024medical, cheng2023sam} such as SAM-Med3D \cite{wang2025sam} aim to transfer knowledge from large-scale natural-image pretraining to medical imaging tasks. Besides, DINOv3 \cite{simeoni2025dinov3} has shown promising performance in medical image segmentation due to enriched deep features. Despite advances in architectural design and pretraining strategies, the impact of dataset-specific properties on model training remains under-explored, particularly voxel spacing and resampling protocols.

\subsection{Volumetric Image Super-resolution}
This task is equivalent to medical image upsampling, which is much more challenging compared with image downsampling. Specifically, Peng et. al pioneeringly introduce a Spatial-Aware Interpolation NeTwork (SAINT) \cite{peng2020saint}, which utilizes voxel spacing information to provide desirable details. TVSRN \cite{yu2022rplhr} adopts a pure Transformer network for volumetric super-resolution, achieving a better trade-off between image quality, the number of parameters, and inference time. ArSSR \cite{wu2022arbitrary} incorporates the implicit voxel function via deep neural networks, to realize the arbitrary upsampling-rate reconstruction of high-resolution images from any low-resolution images. CycleINR \cite{fang2024cycleinr} further enhances the grid sampling with local attention and mitigates over-smoothing by integrating the cycle consistent loss. DP-INR \cite{lyu2025diffusion} effectively combines the prior information learning capabilities of diffusion models \cite{ho2020denoising} with the global information integration capacity of spatio-temporal implicit attention. More recently, arbitrary-scale paradigms such as AnySR \cite{zhan2024anysr} extend reconstruction to any-scale settings, while progressive implicit refinement for anisotropic MRI \cite{zhang2026progressive} models continuous resolution transitions. JOANet \cite{qiu2025joanet} further couples super-resolution pre-processing with segmentation via joint optimization, yet operates on 2D slices without addressing volumetric voxel-spacing consistency. However, the aforementioned methods fail to explicitly model the inter-slice correlation, leading to anatomical non-smoothness along the inter-slice direction and causing structural discontinuities.

\subsection{Spacing-oriented Segmentation}
SynthSeg \cite{billot2023synthseg} is the first segmentation CNN robust against changes in contrast and resolution. SynthSeg is trained with synthetic data sampled from a generative model conditioned on segmentations. Crucially, a domain randomisation strategy is adopted by randomising the contrast and resolution of the synthetic training data. Consequently, SynthSeg can segment real scans from a wide range of target domains without retraining or fine-tuning. HyperSpace \cite{joutard2024hyperspace} proposes to condition segmentation models on the voxel spacing using hypernetworks. This approach allows processing images at their native resolutions or at resolutions adjusted to the hardware and time constraints at inference time. However, these two methods do not consider the task interaction between image segmentation and medical image resampling.

 \begin{figure}[!t]
\centerline{\includegraphics[width=1.0\linewidth]{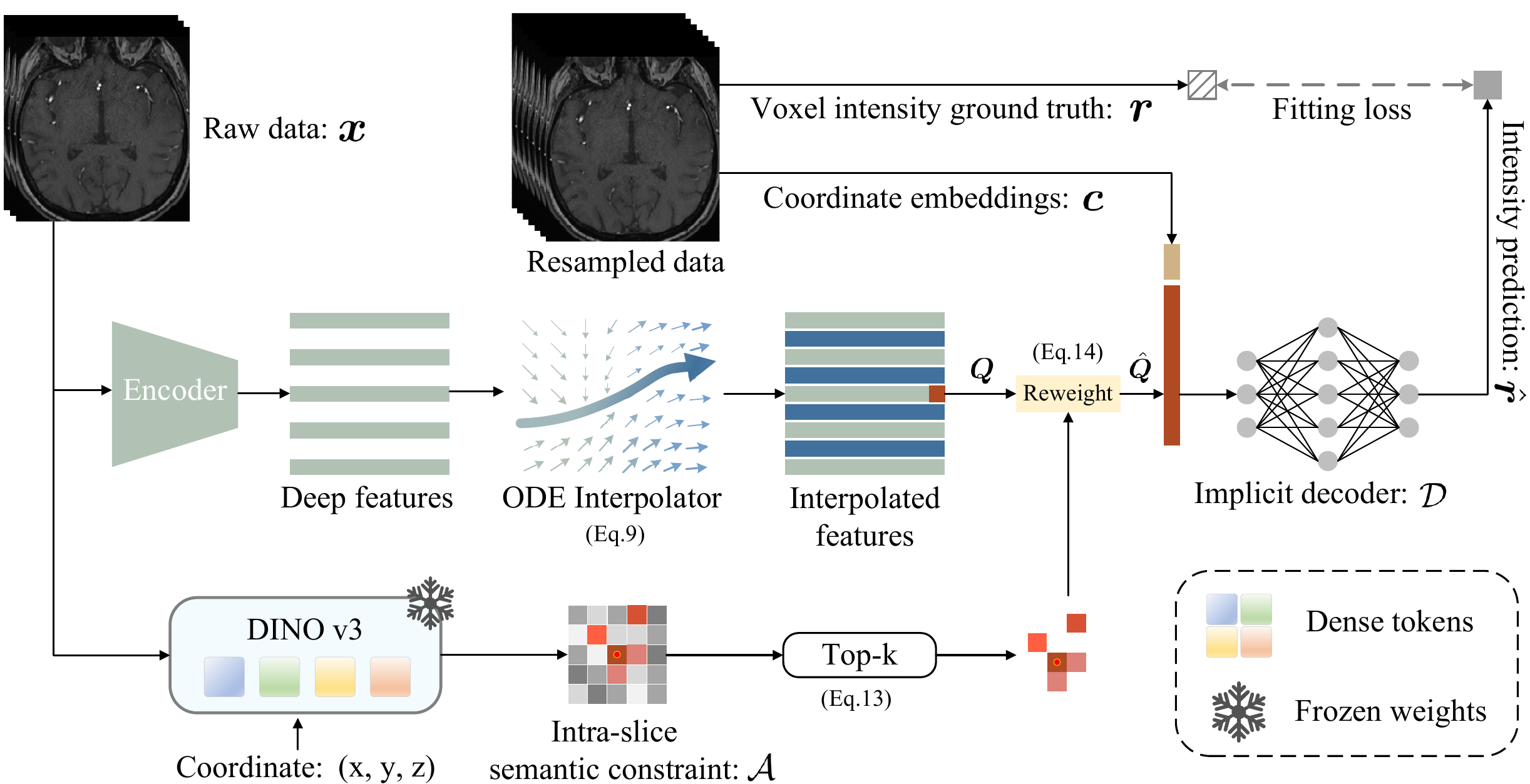}}
  \caption{The overall pipeline of Consispace. The ODE-based interpolator models inter-slice dynamics of anatomical structures in the axial direction. Besides, a pretrained vision model is introduced to extract intra-slice contextual correlations, which are incorporated to reweight interpolated features. Finally, an implicit neural network is proposed to aggregate both inter-slice and intra-slice anatomical and semantic constraints.
  }
  \label{consispace}
\end{figure}

\section{Methodology}
\subsection{Problem Formulation and Whole Pipeline}
The ultimate objective is to achieve promising segmentation performance on volumetric datasets with diverse voxel spacing. To achieve that, we need to optimize the following objective function:
\begin{eqnarray}
& \tilde{\mathcal{S}} = \mathop{\arg\min} \limits_{\mathcal{\theta}} \, \mathcal{L}_{s} (\mathcal{S}_{\theta}(\bm{X}), \bm{Y}),
\label{segmentation target}
\end{eqnarray}
where $\theta$ means the parameters of segmentation networks $\mathcal{S}_{\theta}$, $\bm{X}$ and $\bm{Y}$ refer to the image and mask sets. $\mathcal{L}_{s}$ corresponds to the segmentation loss function, which provides the optimal solution $\tilde{\mathcal{S}}$ in the parameter space. However, as Hyperspace \cite{joutard2024hyperspace} and other literature \cite{isensee2021nnu, kondrateva2022we, seoni2024all} claim, unified voxel spacing can alleviate the inconsistency of physical spacing, thereby boosting anatomical representations of segmentation models. Thus, we design an image resampling framework to enforce a harmonized voxel spacing before feeding the training and testing datasets to the segmentation network. The specific target is as follows:
\begin{eqnarray}
& \mathcal{R}^{*} = \mathop{\arg\min} \limits_{\mathcal{\phi}} \, \mathcal{L}_{r} (\mathcal{R}_{\phi}(\bm{X}_{l}), \bm{X}_{h}),
\label{resampling target}
\end{eqnarray}
where $\phi$ means the parameters of resampling networks $\mathcal{R}_{\phi}$. Since the downsampling operation is generally less challenging, we mainly focus on the upsampling operation. Thus, low-resolution and high-resolution image pairs $\bm{X}_{l}$, $\bm{X}_{h}$ are constructed to model the upsampling process. The reconstruction loss $\mathcal{L}_{r}$ helps generate the optimal reconstruction network $\mathcal{R}^{*}$. Therefore, the final segmentation network $\mathcal{S}^{*}$ should satisfy:
\begin{eqnarray}
& \mathcal{S}^{*} = \mathop{\arg\min} \limits_{\mathcal{\theta}} \, \mathcal{L}_{s} (\mathcal{S}_{\theta}(\mathcal{R}^{*}(\bm{X})), \bm{Y}).
\label{final target}
\end{eqnarray}

In this work, any U-shaped encoder-decoder architecture can be adopted to represent $\mathcal{S}^{*}$. While for the medical image resampling, we choose an implicit neural network \cite{wu2022arbitrary} to achieve image downsampling and upsampling in random scales. This pipeline is aimed at Consistent voxel spacing, termed \textbf{Consispace}. An overview of Consispace is shown in Fig.\ref{consispace}.

\subsection{ODE-based Inter-slice Dynamics}
In this work, volumetric data is regarded as an ordered sequence of 2D axial slices, $\bm{x}=[\bm{x}^{(1)},\ldots,\bm{x}^{(D')}]$. Each slice
$\bm{x}^{(i)}$ belongs to $\mathbb{R}^{H\times W}$. Since adjacent slices within a volume typically exhibit gradual anatomical variations, we treat the slice index $i$ as a pseudo-time variable to model inter-slice transitions. This perspective allows anatomical representations to evolve in a continuous latent space, providing an anatomical regularization mechanism for volumetric resampling beyond voxel-wise interpolation. Specifically, we assume that the target 3D anatomy can be represented by a time-dependent latent field $\bm{Q}(t)\in\mathbb{R}^{D \times H \times W \times d}$, whose evolution is governed by an ODE-inspired dynamics \cite{beg2005computing}:
\begin{equation}
\frac{\partial \bm{Q}}{\partial t}=\bm{v}(\bm{Q},t).
\label{eq:ode}
\end{equation}
Here, $\bm{Q}(t)$ encodes slice-wise anatomical representations, and the time-varying velocity field $\bm{v}$ characterizes their smooth evolution along the axial direction.

The continuous formulation above serves as a modeling prior for inter-slice anatomical evolution, while in practice the computation is performed at discrete slice index. Starting from an initial condition $\bm{Q}^{(0)}$, we discretize the trajectory of $\bm{Q}(t)$ with a step size $\Delta t$, obtaining latent states $\bm{Q}^{(i\Delta t)}$ for $i=1,\ldots,D$. The dynamics in Eq.~\eqref{eq:ode} are then numerically approximated by the Euler method:
\begin{equation}
\bm{Q}^{(i)}=\bm{Q}^{(i-1)}+\Delta t\, \bm{v}^{(i-1)}\!\left(\bm{Q}^{(i-1)}\right),
\quad i=1,\ldots,D.
\label{eq:euler}
\end{equation}
$\Delta t$ is empirically simplified as 1 \cite{lipman2022flow}. Since $\bm{Q}$ represents the 2D anatomical axial slices, we expect that the evolution of
$\bm{Q}$ should be guided by next-slice image information. Therefore, the velocity field should depend on $\bm{x}^{(i)}$. Given an INR-based encoder $\mathcal{E}$ that maps $\bm{x}^{(i)}$ to a latent representation
$\bm{z}^{(i)}$, i.e.,
$\bm{z}^{(i)}=\mathcal{E}(\bm{x}^{(i)})$,
the time-dependent velocity field $\bm{v}^{(i-1)}$ can be written as:
    \begin{equation}
\bm{v}^{(i-1)}\!\left(\bm{Q}^{(i-1)}\right)=f\!\left(\bm{Q}^{(i-1)}, \bm{z}^{(i)}\right).
\label{eq:velocity}
\end{equation}
Here $f$ refers to the functional mapping for calculating velocity fields, transforming $\bm{Q}^{(i-1)}$ into $\bm{Q}^{(i)}$. By incorporating the velocity field $\bm{v}$ into the image resampling architecture,
the implicit decoder $\mathcal{D}$ takes both time-varying representations $\bm{Q}$ and coordinate embeddings $\bm{c}$ as inputs to produce a
resampled image $\hat{\bm{r}}$, i.e.,
$\hat{\bm{r}}=\mathcal{D}(\bm{Q}, \bm{c})$.
We train the implicit neural network so that the prediction
$\hat{\bm{r}}$ is close to the ground-truth image $\bm{r}$.


\subsection{ODE-based Interpolator}
The aforementioned time-varying representations $\bm{Q}$ are generated based on next-slice image latents in an iterative manner. However, since each slice in the volumetric data correlates with both upper and lower slices in the axial direction, single-directional slice progression alone cannot adequately describe the inter-slice dynamics. Therefore, bi-directional ODE constraints are enforced to maintain smoother anatomical variations. The specific formulation is as follows:
\begin{eqnarray}
& \bm{Q}_{\downarrow}^{(i)}=\bm{Q}_{\downarrow}^{(i-1)}+\Delta t\, f_{1}\!\left(\bm{Q}_{\downarrow}^{(i-1)}, \bm{z}^{(i)}\right), \quad \bm{Q}_{\downarrow}^{1}=\bm{z}^{(1)}, \\
& \bm{Q}_{\uparrow}^{(i)}=\bm{Q}_{\uparrow}^{(i+1)}+\Delta t\, f_{2}\!\left(\bm{Q}_{\uparrow}^{(i+1)}, \bm{z}^{(i)}\right),  \quad \bm{Q}_{\uparrow}^{D}=\bm{z}^{(D)},
\label{bi-direction}
\end{eqnarray}
where $\bm{Q}_{\downarrow}$ and $\bm{Q}_{\uparrow}$ correspond to temporal representations in a top-down and bottom-up manner. For the top-down style, the ODE evolution in Eq.\eqref{bi-direction} requires the initial condition $\bm{Q}_{\downarrow}^{(1)}$, which is essential for the iterative update. Since $\bm{Q}_{\downarrow}^{(1)}$ is intended to capture anatomical information of the first slice after the ODE-based interpolator, we directly adopt the latent representation of the first slice $\bm{z}^{(1)}$ to define it. Analogously, $\bm{z}^{(D)}$ is set as the initial state of $\bm{Q}_{\uparrow}^{(D)}$. Through a linear combination of these two elements, we can acquire the time-varying representations $\bm{Q}^{(i)}$ that satisfy:
\begin{eqnarray}
\bm{Q}^{(i)} &=& \underbrace{\frac{\bm{Q}_{\uparrow}^{(i+1)} + \bm{Q}_{\downarrow}^{(i-1)}}{2}}_{\mathrm{\textbf{Average}}} \nonumber\\
&& + \sigma \underbrace{[\, f_{1}(\bm{Q}_{\downarrow}^{(i-1)}, \bm{z}^{(i)}) + f_{2}(\bm{Q}_{\uparrow}^{(i+1)}, \bm{z}^{(i)})]}_{\mathrm{\textbf{Instantaneous\;Velocity\;Field}}},
\label{internal energy function}
\end{eqnarray}
where $\sigma$ is a constant value equal to $\Delta t/2$. $\bm{Q}^{(i)}$ integrates both top-down and bottom-up contextual information, which smooths inter-slice variations and promotes anatomical consistency across slices. More concretely, we decompose $\bm{Q}^{(i)}$ into two terms. As shown in Eq.\eqref{internal energy function}, the first term is the slice-wise average representation. The second term encodes the influences of instantaneous velocity fields, which warp representations from neighboring slices to slice $i$, thus yielding more consistent inter-slice evolution.

Next, we will discuss the implementation of functional mapping $f_{1}, f_{2}$ for modeling instantaneous velocity fields between slices. Take $f_{1}$ as an example. Velocity fields mainly capture inter-slice differences. Thus, we select $\mathcal{I}_{t}$ as the target information, as illustrated by the following formulas:
\begin{gather}
\mathcal{I}_{t}
= \left(\bm{Q}^{(i-1)} - \bm{V}^{(i)}\right),
\label{eq:target_info} \\
S
= \frac{
\bm{Q}^{(i-1)} \cdot \bm{K}^{(i)}
}{
\left\lVert \bm{Q}^{(i-1)} \right\rVert_{1}
\left\lVert \bm{K}^{(i)} \right\rVert_{1}
},
\label{eq:similarity} \\
f_{1}\left(\bm{Q}^{(i-1)}, \bm{z}^{(i)}\right)
= \operatorname{Softmax}\left(1-S\right)\mathcal{I}_{t}.
\label{eq:cross_attn}
\end{gather}
Specifically, $\bm{K}^{(i)}, \bm{V}^{(i)}$ refers to the key and value transform of $\bm{z}^{(i)}$. Furthermore, it is significant to determine the attention score within $\mathcal{I}_{t}$. Indeed, attention should be attenuated for token pairs with high similarity and strengthened for pairs with pronounced differences. Motivated by this point, we adopt the reversed cosine similarity metric. As shown in Eq.\eqref{eq:cross_attn}, $S\in\mathbb{R}^{HW \times HW}$ denotes the pairwise cosine-similarity matrix between $\bm{Q}^{(i-1)}$ and $\bm{K}^{(i)}$. We then compute the discrepancy matrix as $1 - S$, where larger values indicate greater dissimilarity between tokens. Notably, this matrix is already properly scaled and can be directly input to the Softmax operation. Finally, the normalized weight is combined with target information $\mathcal{I}_{t}$ to better model instantaneous velocity fields.

\subsection{Intra-slice Constraint}
The ODE-based interpolator can promote segmentation through a unified voxel spacing, as demonstrated by nn-UNet \cite{isensee2021nnu, joutard2024hyperspace}. However, we are encouraged to explore the bi-directional interaction between image resampling and semantic segmentation in this work. Thus, resampling performance is to be improved by introducing additional semantic guidance, which is the key to addressing the ill-posed upsampling problem. 

Meanwhile, the proposed interpolator only exploits the inter-slice dynamics of anatomical structures, neglecting the intra-slice contextual correlations. Thus, we leverage a pretrained vision model to produce high-quality dense representations for 2D slices, which construct a contextual correlation map for the specifically queried location. Specifically, given the dense feature field of a slice $\bm{Q}^{(i)}$, we compute the correlation between the feature vector at the target point $p$ and all other feature vectors using a cosine-similarity metric, yielding a dense map $\mathcal{A}$ that reflects semantic constraints within the slice. This contextual correlation map enables the identification of a semantically coherent neighborhood around the target point $p$. The detailed process can be formulated as follows:
\begin{align}
\mathcal{N}_k(p) &= \operatorname*{Top-k}_{j \in \mathcal{P}} \, A(j,p), \\
\hat{\bm{Q}}_{p}^{(i)} &= \displaystyle \sum_{j \in \mathcal{N}_k(p)} \bm{Q}^{(i)}_{j} \mathcal{A}(j, p),
\end{align}
\label{intra-slice}
where $\mathcal{P}$ is the local window centered at point $p$, $\mathcal{N}_k(p)$ contains points with the top-k highest similarities. By selecting the top-k most correlated locations $\mathcal{N}_k(p)$, we then perform a weighted aggregation to refine the interpolated features $\bm{Q}^{(i)}_{p}$ as enhanced features $\hat{\bm{Q}}^{(i)}_{p}$, where the weights $\mathcal{A}(j, p)$ are derived from the normalized similarity scores. By propagating information from semantically consistent neighbors, the proposed strategy promotes class-wise semantic consistency of intra-slice anatomical structures, which is beneficial for the downstream segmentation task. 

\section{Experiments}
\subsection{Datasets}
\label{datasets}
\noindent\textbf{TopCow 2024~\cite{yang2025benchmarking}:}
This public dataset contains 125 brain-vessel MRI image/mask pairs, with voxel spacing ranging from $[0.30, 0.39]\times[0.30, 0.39]\times[0.50, 0.80]\ \mathrm{mm}^3$. We first hold out 30 cases as the test set for the segmentation task, which reflects the real inference scenario without synthetic spacing perturbation. For each of the 125 cases, we further generate a large-spacing counterpart by downsampling the image along the axial direction with a random scaling factor sampled from $(1.0, 3.0)$. Moderate Gaussian blur and Gaussian noise are also introduced to better simulate the real data acquisition process. In this way, each original image and its downsampled version form a large-spacing/small-spacing pair for the resampling task.

For the segmentation task, the 30 held-out real cases are used only for testing. The remaining 95 real cases and their corresponding downsampled counterparts form 190 image/mask pairs, which are further split into 150 training cases and 40 validation cases. For the resampling task, all 125 large-spacing/small-spacing pairs are split at the case level into 80 training pairs, 15 validation pairs, and 30 testing pairs. The 30 testing pairs correspond to the same held-out cases used for segmentation testing, ensuring that no case-level overlap exists between training and testing. The ground-truth segmentation includes 14 classes, where label 0 denotes the background and labels $1\mbox{-}13$ correspond to 13 distinct brain vessels.


\noindent\textbf{BraTS 2020~\cite{menze2014multimodal, bakas2018identifying}:}
BraTS 2020 contains 369 brain MRI scans. Each case has a volume size of $155 \times 240 \times 240$ and an isotropic voxel spacing of $1 \times 1 \times 1\ \mathrm{mm}^3$. To evaluate the influence of voxel-spacing inconsistency, we adopt the same axial downsampling and Gaussian operations as used for TopCow 2024. Following this protocol, 69 image/mask pairs and 69 corresponding small-spacing/large-spacing pairs are held out for segmentation and resampling inference. For the resampling task, 240 and 60 data pairs are used for training and validation. For the segmentation task, 600 image/mask pairs are split into 480 training pairs and 120 validation pairs. It should be noted that no original case or its downsampled counterpart appears across training and testing splits. The ground-truth segmentation includes four labels: 0 for background, 1 for non-enhancing tumor core, 2 for peritumoral edema, and 4 for GD-enhancing tumor. Due to ambiguous tumor boundaries and highly irregular morphologies, BraTS provides a challenging benchmark for evaluating both resampling fidelity and downstream 3D segmentation performance.

\noindent\textbf{SPIDER \cite{van2024lumbar}:} SPIDER is a public lumbar-spine MRI dataset containing 447 scans, including 41 small-spacing cases. Following the same experimental protocol, we construct paired data by resampling high-resolution volumes to coarser axial spacings with random scaling factors twice. 10 image/mask pairs and 10 corresponding small-spacing/large-spacing pairs are held out for segmentation and resampling inference. For the resampling task, 31 data pairs are used for training. For the segmentation task, 93 image/mask pairs are split into 73 training pairs and 20 validation pairs. The ground-truth segmentation contains 20 classes: class $0$ denotes the background, classes $1\mbox{-}9$ correspond to nine lumbar vertebrae, class $10$ denotes the spinal canal, and classes $11\mbox{-}19$ correspond to intervertebral discs.

\subsection{Experimental Settings}
\noindent \textbf{Implementation Details.} For training medical image resampling frameworks, as described in Sec. \ref{datasets}, when constructing the training and validation sets, we only apply linear resampling from smaller slice spacing to larger slice spacing to generate paired high-/low-resolution volumes, instead of resampling to an even finer spacing. This choice is motivated by the fact that resampling to smaller spacing may introduce unreliable details and thus lower imaging quality. MedNeXt \cite{roy2023mednext} is selected as the encoder of Consispace, which is trained with the reconstruction loss function. Here we adopt a combination of MSE loss and SSIM loss, and the weights are set as $1.0$ and $1.0$. All resampling models are trained using the AdamW \cite{loshchilov2017decoupled} optimizer with the linear warm-up strategy for the first 25 epochs. The learning rate, batch size and training epoch are equal to $1e\mbox{-}4$, $4$, and $500$. For the intra-slice semantic constraint, we instantiate the pretrained vision model with DINOv3 \cite{simeoni2025dinov3} to extract dense slice features, and the effect of alternative feature backbones is studied in the ablation experiments.

For training segmentation networks, we use nnU-Net \cite{isensee2021nnu}, Swin Transformer \cite{tang2022self}, and MedSAM-2 \cite{zhu2024medical} as baselines to assess the impact of voxel spacing unification. All experiments are trained with the AdamW optimizer, with a warm-up cosine scheduler for the first 50 epochs. The learning rate, batch size, and training epoch are set as $5e\mbox{-}4$, $2$, and $1000$. The segmentation loss function is the linear combination of Dice loss and Cross entropy loss, with weights equal to $1.0$ and $1.0$. For data augmentation, we adopt strategies of random rotation between [$\mbox{-}15^{\circ}, 15^{\circ}$], random flipping along the XOZ/YOZ plane, and random intensity scaling between $[0.9, 1.1]$. All experiments are implemented based on Pytorch 2.1.0 and 2 NVIDIA RTX 5090 GPUs.

\begin{table*}[!t]
  \begin{center}
  \caption{(a) Quantitative segmentation performance comparisons on different settings of voxel spacing. (b) Comparison with other methods for enhancing the voxel spacing. Two strong segmentation baselines are adopted, including nnU-Net and Swin UNETR. Dice score and HD95 are incorporated as the average metrics for all segmentation classes.}
  \label{tab segmentation}
  \resizebox{0.94\linewidth}{!}{
  \begin{tabular}{llcccccc}  
\toprule  
  \multicolumn{8}{c}{(a) Different settings of voxel spacing} \\
\midrule
  \multirow{2}*{Model} &  \multirow{2}*{Setting} & \multicolumn{2}{c}{SPIDER} & \multicolumn{2}{c}{TopCow24}  & \multicolumn{2}{c}{BraTS20} \\  
  \cmidrule(r){3-4}  \cmidrule(r){5-6}  \cmidrule(r){7-8}
 &  & Dice (\%) $\uparrow$ & HD95 (mm) $\downarrow$ & Dice (\%) $\uparrow$ & HD95 (mm)$\downarrow$ & Dice (\%) $\uparrow$ & HD95 (mm) $\downarrow$ \\
\midrule
 &  Original spacing & 81.64 & 14.19 & 85.59 & 4.16 & 85.13 & 7.06    \\
nnU-Net \cite{isensee2021nnu} &  Uniform spacing (Linear) & 89.13 & 4.87 & 88.42 & 2.72  & 86.06 & 6.31   \\
 &  Uniform spacing (Ours) & \textbf{90.56} & \textbf{3.48} & \textbf{90.83} & \textbf{1.88} & \textbf{87.20} & \textbf{5.39}   \\
\midrule 
 &  Original spacing & 81.91 & 15.14 & 85.72 & 3.98 & 84.20 & 9.84    \\
Swin UNETR \cite{tang2022self} &  Uniform spacing (Linear) & 87.28 & 10.56 & 87.43 & 3.19 & 84.85 & 8.40   \\
 &  Uniform spacing (Ours) & \textbf{88.32} & \textbf{6.84} & \textbf{88.76} & \textbf{2.56} & \textbf{85.72} & \textbf{6.97}   \\
\midrule
  \multicolumn{8}{c}{(b) Voxel-spacing-oriented methods} \\
\midrule
 &  Synthseg \cite{billot2023synthseg} & 87.49 & 6.30 & 87.17 & 3.68 & 85.79 & 6.05  \\
nnU-Net \cite{isensee2021nnu} &  HyperSpace \cite{joutard2024hyperspace}  & 88.79 & 5.73 & 89.24 & 2.37 & 86.11 & 5.80   \\
 &  Ours  & \textbf{90.56} & \textbf{3.48} & \textbf{90.83} & \textbf{1.88}  & \textbf{87.20} & \textbf{5.39}  \\
\bottomrule
  \end{tabular}}
  \end{center}
\end{table*}

\begin{figure*}[!t]
\vspace{-4mm}
\centerline{\includegraphics[width=0.8\linewidth]
{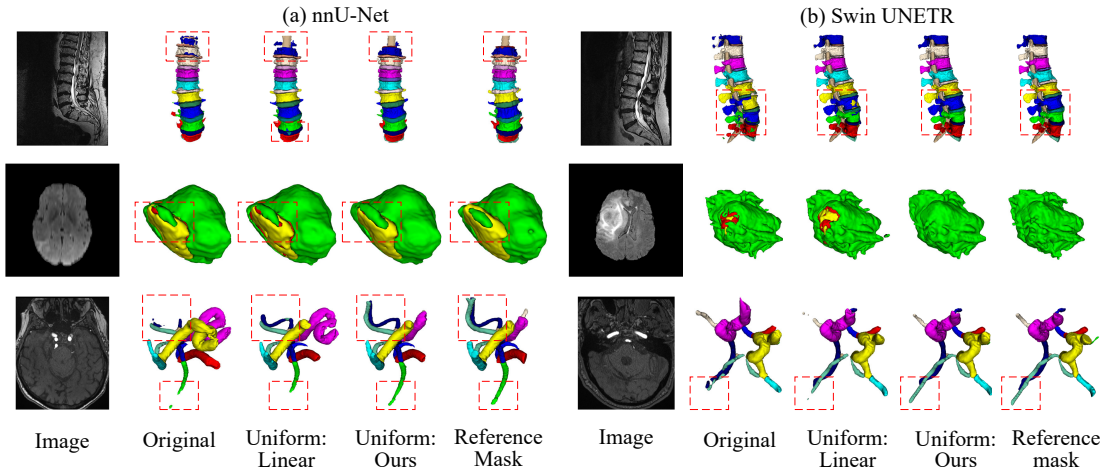}}
  \caption{The performance comparison between linear interpolation and the proposed Consispace on medical image segmentation.
  }
  \label{segmentation}
\end{figure*}

\noindent \textbf{Evaluation Metrics.} For the image resampling task, we select evaluation metrics including Peak Signal-to-Noise Ratio (PSNR) \cite{hore2010image}, Structural Similarity (SSIM) \cite{wang2004image}, Normalized Mean Square Error (NMSE) \cite{kim2024data}, Learned Perceptual Image Patch Similarity (LPIPS) \cite{zhang2018unreasonable}. Compared to PSNR and SSIM, which are traditional pixel-wise reconstruction metrics, LPIPS evaluates perceptual similarities from the perspective of human vision \cite{jain2024video, you2025fb}. Therefore, this metric is more feasible and favored in clinical practice. For the segmentation task, the Dice score \cite{milletari2016v} is adopted as the overlap-based metric, and $95\%$ Hausdorff distance (HD95) \cite{karimi2019reducing} is used to measure the performance of boundary delineation.

\subsection{Experimental Results}
\label{4.3}
\noindent \textbf{\textit{Comparative analysis on the types of voxel spacing.}} As we mentioned in the Introduction, there are three settings related to the voxel spacing, including:
\begin{itemize}
    \item Original spacing: All images and masks are trained in a segmentation network, maintaining the original voxel spacing at training and inference time.
    \item Uniform spacing (Linear): All training images are resampled to the median voxel spacing using linear interpolation, while the corresponding masks are resampled using nearest-neighbor interpolation. The resampled data are then used to train the segmentation network. During inference, images are first resampled to the unified spacing, and model predictions are subsequently mapped back to the original spacing via inverse resampling.
    \item Uniform spacing (Ours): Compared with the setting of Uniform spacing (Linear), we aim to resample the volumetric images with the proposed Consispace framework in the training and inference stages. The other steps remain the same.
\end{itemize}

We adopt the classical nnU-Net \cite{isensee2021nnu} and Swin UNETR \cite{tang2022self} as the segmentation baselines to evaluate quantitative and qualitative performance. As revealed in Table \ref{tab segmentation}(a), after unifying the voxel spacing of SPIDER datasets in a vanilla linear way, there is $7.49\%\uparrow$ Dice score and $9.32mm\downarrow$ HD95 based on nnU-Net. This result demonstrates the effectiveness of unifying voxel spacing in the medical image segmentation task, as has been claimed in previous literature \cite{isensee2021nnu, joutard2024hyperspace}. 
\begin{figure*}[!t]
\vspace{-4mm}
\centerline{\includegraphics[width=0.8\linewidth]{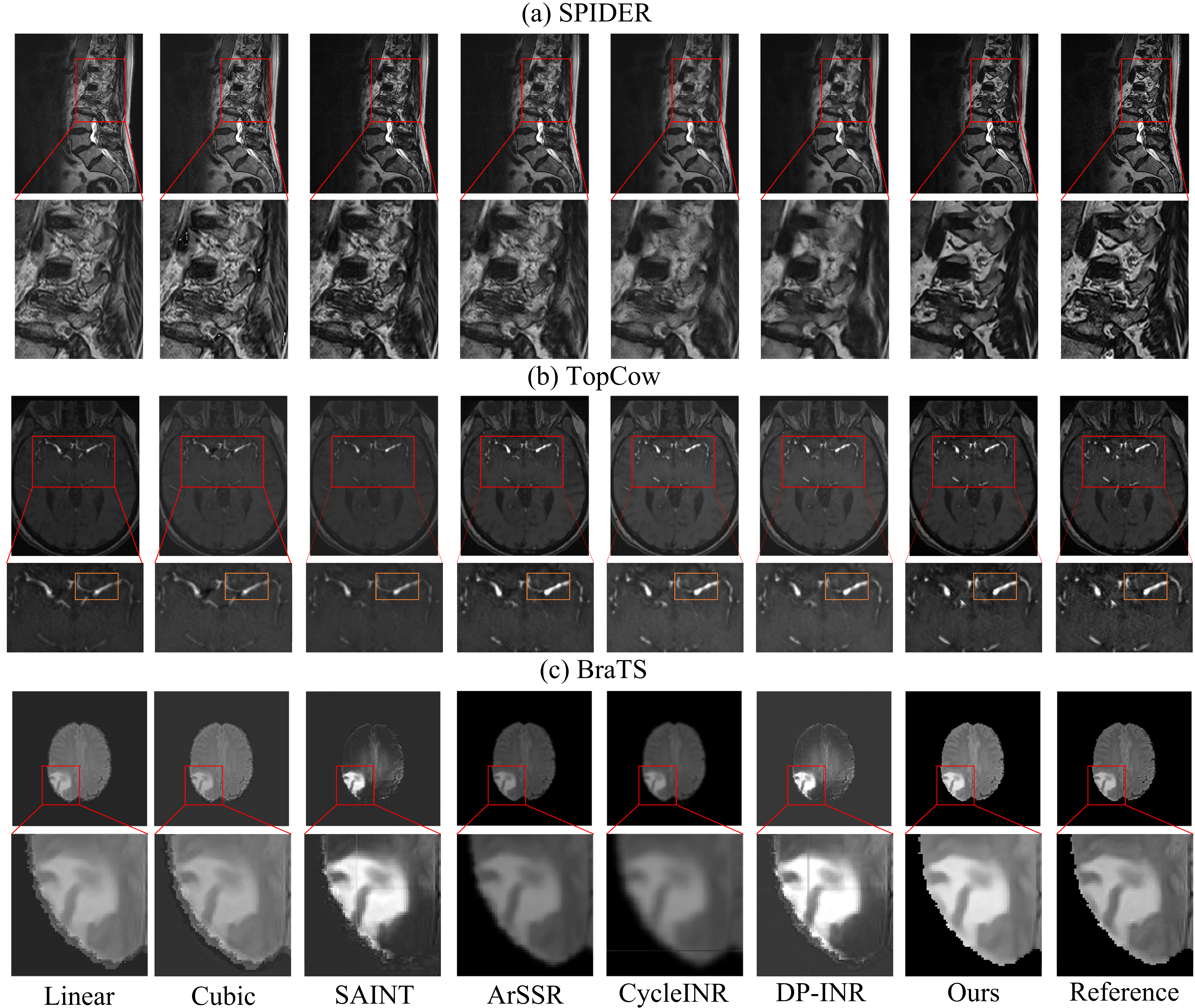}}
  \caption{The image resampling performance comparison between the proposed Consispace and other resampling approaches.
  }
  \label{reconstruction}
\end{figure*}

\begin{table*}[!t]
  \begin{center}
\captionsetup{aboveskip=1pt, belowskip=0pt}
  \caption{Comparison with other methods on volumetric image resampling when evaluated on SPIDER, TopCow24 and BraTS20. Evaluation metrics include PSNR (dB), SSIM, NMSE, and LPIPS (The best results are highlighted in bold, and the second-best results are underlined).}
  \label{tab resampling}
  \resizebox{1.0\linewidth}{!}{
  \begin{tabular}{lcccccccccccc}  
\toprule  
  \multirow{2}*{Method} & \multicolumn{4}{c}{SPIDER} & \multicolumn{4}{c}{TopCow24} & \multicolumn{4}{c}{BraTS20} \\  
  \cmidrule(r){2-5}  \cmidrule(r){6-9}   \cmidrule(r){10-13}
 & PSNR (dB) $\uparrow$ & SSIM $\uparrow$ & NMSE $\downarrow$ & LPIPS $\downarrow$ & PSNR (dB) $\uparrow$ & SSIM $\uparrow$ & NMSE $\downarrow$ & LPIPS $\downarrow$ & PSNR (dB) $\uparrow$ & SSIM $\uparrow$ & NMSE $\downarrow$ & LPIPS $\downarrow$  \\
\midrule
 Linear   & 23.15 & 0.563 & 0.084 & 2.228 &  30.89 & 0.843 & 0.042  &  1.429  & 19.15 & 0.567 & 0.333 & 3.006  \\
 Cubic  & 19.50 & 0.568 & 0.156 & 2.949 & 25.61 & 0.745 & 0.108 & 1.817  & 16.95  & 0.457  &  0.441 &  3.694   \\
 SAINT \cite{peng2020saint}   & 25.55 & 0.717 & 0.082 & 1.912 & 32.64  & 0.872 & 0.034 & 1.134 & 23.45 & 0.876 & 0.216 & 2.542 \\
 TVSRN \cite{yu2022rplhr}   & 26.87 & 0.774 & 0.061 & 1.746 & 32.93 & 0.879 & 0.037 & 1.179 & 24.68 & 0.889 & 0.170 & 2.312 \\
 ArSSR \cite{wu2022arbitrary}  & 26.32 & 0.763 & 0.069 & 1.714 & 33.08 & 0.880 & 0.030 & 1.052  & 24.89 & 0.903 & 0.141 &  2.347 \\
 CycleINR \cite{fang2024cycleinr}   & 27.20 & 0.802 & 0.055 & \underline{1.651} & \underline{34.62} &  \underline{0.905} &  \textbf{0.017} & \underline{0.891} & \underline{25.06} & \underline{0.917} & 0.101 & \textbf{1.860} \\
 DP-INR \cite{lyu2025diffusion}  & \underline{27.54} & \underline{0.813} & \underline{0.052} & 1.677 & 34.29  & 0.897 & 0.024 & 0.922 & 24.74 & 0.911 & \underline{0.092} & 2.129  \\
 Consispace  & \textbf{28.31} & \textbf{0.849} & \textbf{0.039} & \textbf{1.418} & \textbf{35.09} & \textbf{0.916} & \underline{0.019} &  \textbf{0.768} & \textbf{25.48} & \textbf{0.925} & \textbf{0.078} & \underline{1.896} \\
\bottomrule 
  \end{tabular}}
  \end{center}
\end{table*}

Since linear interpolation is prone to inter-slice non-smoothness, it may result in structural discontinuities. It also leads to distorted boundaries between different anatomies, causing poor semantic consistency within each segmentation class. Thus, we propose Consispace to achieve more effective image resampling. As shown in Table \ref{tab segmentation}(a), by replacing linear interpolation with Consispace, there are $1.43\%\uparrow$, $2.41\%\uparrow$, and $1.14\%\uparrow$ Dice score increases on SPIDER, TopCow24 and BraTS20 based on nnU-Net. Besides, we also provide segmentation visualizations among three settings of voxel spacing. As illustrated in Fig.\ref{segmentation}, resampled images generated by Consispace achieve fine-grained segmentation results, which contain brain vessels with better connectivity, brain tumors with precise boundaries, and vertebrae with intra-class semantic consistency. 



\noindent \textbf{\textit{Comparison with methods on volumetric image resampling.}} We investigate the performance analysis of volumetric image resampling. Based on the quantitative results in Table \ref{tab resampling}, our method consistently achieves the best performance on SPIDER, TopCow24, and BraTS20 across four reconstruction metrics. On SPIDER, we obtain 28.31dB PSNR and 0.849 SSIM, while reducing NMSE and LPIPS to 0.039 and 1.418, respectively, outperforming all competing approaches. A similar trend is observed on TopCow24 and BraTS20. Besides, we present qualitative image resampling results. As revealed in Fig.\ref{reconstruction}, Consispace improves both inter-slice and intra-slice fidelity by better preserving fine-grained anatomical textures and boundaries. Specifically, there are fewer generated fake vessels in the upsampled TopCow data, and more accurate vertebral boundaries in SPIDER data.

\begin{table*}[!t]
  \begin{center}
  \caption{Structural ablation on the core components of Consispace, including ODE-based anatomical constraints and semantic consistency constraints derived from DINOv3. We also consider the efficacy of bidirectional anatomical information and reversed attention inside the ODE interpolator.}
  \label{tab ablation}
  \resizebox{0.89\linewidth}{!}{
  \begin{tabular}{lcccccccc}  
\toprule 
  \multirow{2}*{Settings} & \multicolumn{4}{c}{SPIDER} & \multicolumn{4}{c}{TopCow24}  \\  
  \cmidrule(r){2-5}  \cmidrule(r){6-9}  
 & PSNR (dB) $\uparrow$ & SSIM $\uparrow$ & NMSE $\downarrow$ & LPIPS $\downarrow$ & PSNR (dB) $\uparrow$ & SSIM $\uparrow$ & NMSE $\downarrow$ & LPIPS $\downarrow$ \\
\midrule
 Consispace   & \textbf{28.31} & \textbf{0.849} & \textbf{0.039} & \textbf{1.418} & \textbf{35.09} & \textbf{0.916} & \textbf{0.017} &  \textbf{0.768} \\
 w/o DINOv3  & 27.89 & 0.828 & 0.048 & 1.605 & 34.76  & 0.905 & 0.025 & 0.847   \\
 w/o ODE constraints  & 27.11 & 0.803 & 0.059 & 1.634 & 34.23 & 0.891 & 0.026 &  0.955  \\
 w/o both  & 26.97 & 0.796 & 0.057 & 1.698 & 33.10 & 0.880 & 0.035 &  1.119  \\
\midrule  
 Unidirectional-ODE  & 27.97 & 0.832 & 0.052 & 1.613 & 34.28 & 0.889 & 0.029 &  0.923  \\
 Bidirectional-ODE (ours)   & \textbf{28.31} & \textbf{0.849} & \textbf{0.039} & \textbf{1.418} & \textbf{35.09} & \textbf{0.916} & \textbf{0.017} &  \textbf{0.768} \\
\midrule 
 Vanilla attention  & 27.44 & 0.803 & 0.056 &  1.620 & 34.56  & 0.903  & 0.025 &  0.891  \\
 Reversed attention (ours)   & \textbf{28.31} & \textbf{0.849} & \textbf{0.039} & \textbf{1.418} & \textbf{35.09} & \textbf{0.916} & \textbf{0.017} &  \textbf{0.768} \\
\bottomrule  
  \end{tabular}}
  \end{center}
\end{table*}

\noindent \textbf{\textit{Comparison with voxel-spacing-oriented approaches.}} Table \ref{tab segmentation}(b) demonstrates that our method yields more significant improvements in downstream segmentation compared with prior voxel-spacing-oriented methods across three datasets. In contrast to Synthseg \cite{billot2023synthseg} and Hyperspace \cite{joutard2024hyperspace}, we achieve the highest Dice score and the lowest HD95 on testing datasets of SPIDER ($90.56\%/3.48mm$), TopCow24 ($90.83\%/1.88mm$), and BraTS20 ($87.20\%/5.39mm$). Indeed, DINOv3 is incorporated to extract high-quality semantic similarities. These semantic cues are injected by reweighting interpolated features generated by the ODE interpolator, boosting the intra-class semantic consistency within anatomical structures. Thus, the proposed framework effectively bridges the upstream resampling task and the downstream segmentation objective, rather than treating them as two isolated tasks. This contributes to the superior downstream segmentation performance of our method compared with other spacing-oriented approaches.

\begin{table}[!t]
\centering
\caption{Model efficiency comparison on the resampling task when evaluated on the SPIDER dataset (The inference time refers to the average speed per volume).}
\label{tab:efficiency}
\resizebox{0.95\columnwidth}{!}{
\begin{tabular}{lcccc}
\toprule
Metric & Linear & TVSRN  & CycleINR &  Ours \\
\midrule
Parameters (M)  & - &  4.05 & 7.18 & 2.93 \\
Inference time (s)  & 0.23  & 8.32 & 20.10 & 14.47 \\
GPU memory (GB) & 0.35 &  7.34 & 12.09 & 10.18 \\
FLOPs (G) & 4.87  & 690.50 & 382.63 & 237.89 \\
\bottomrule
\end{tabular}
}
\vspace{-4mm}
\end{table}

\noindent \textbf{\textit{Model efficiency.}}
We further evaluate the computational efficiency of different resampling methods on the SPIDER dataset, as reported in Table~\ref{tab:efficiency}. As expected, linear interpolation has the lowest computational cost, since it is a non-learnable operation without anatomical or semantic modeling. Among learning-based methods, our method uses the fewest trainable parameters, i.e., 2.93M, compared with 4.05M for TVSRN and 7.18M for CycleINR, where the frozen pretrained vision model weights are excluded from parameter counting. In addition, Consispace requires 237.89G FLOPs, which is substantially lower than TVSRN and CycleINR. These results indicate that Consispace achieves a favorable trade-off between reconstruction performance and computational efficiency.

\begin{figure*}[!t]
\centerline{\includegraphics[width=1.0\linewidth]{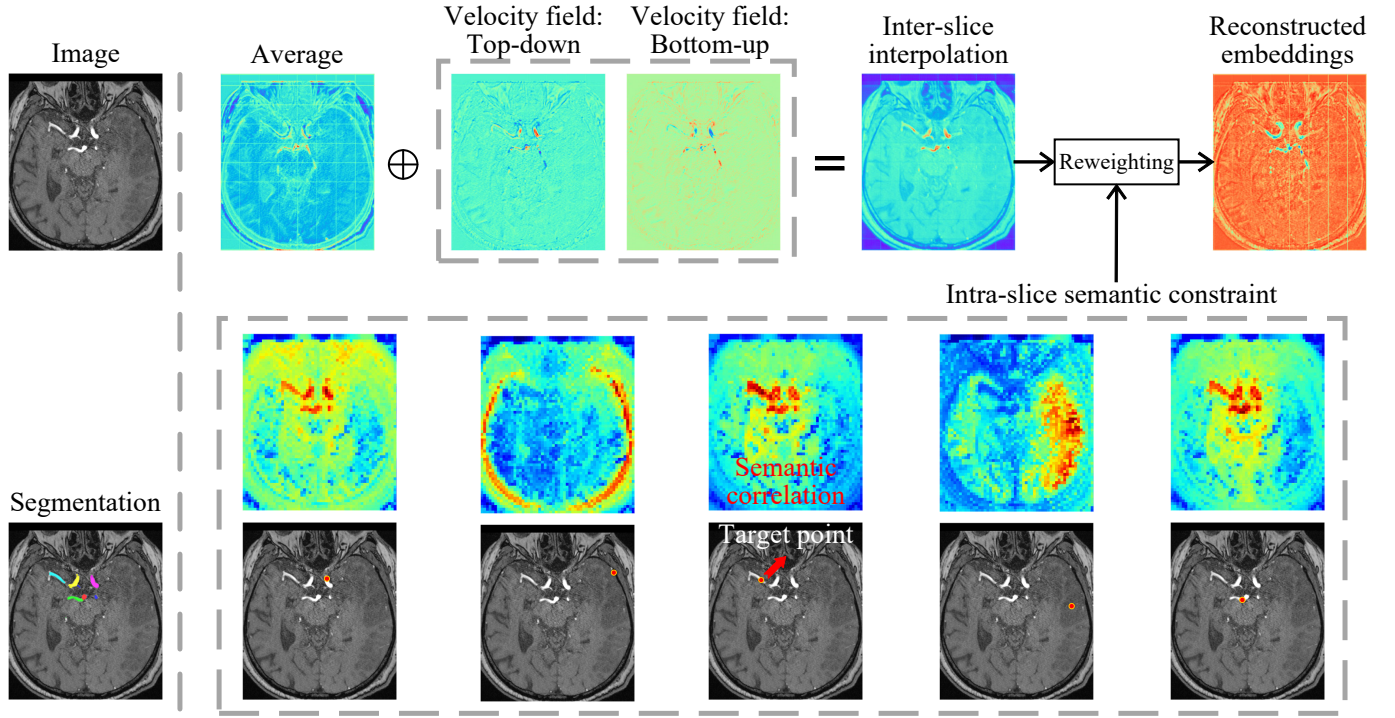}}
\captionsetup{aboveskip=0pt, belowskip=1pt}
  \caption{The detailed feature resampling process in the proposed Consispace.
  }
  \label{feature evolution}
\end{figure*}

\subsection{Ablation Study}

\noindent \textbf{\textit{Structural ablation on Consispace.}}
We first conduct a structural ablation study to evaluate the significance of two key components in Consispace, i.e., the ODE-based anatomical constraints and the intra-slice semantic constraints derived from DINOv3. As reported in Table~\ref{tab ablation}, removing either component leads to consistent degradation across all reconstruction metrics on both SPIDER and TopCow24, demonstrating the necessity of both inter-slice anatomical modeling and intra-slice semantic guidance.

Specifically, removing the DINOv3-guided semantic constraint decreases the PSNR from 28.31 dB to 27.89 dB on SPIDER and from 35.09 dB to 34.76 dB on TopCow24, accompanied by increased LPIPS values. This indicates that DINOv3-based semantic correlations provide effective contextual priors for preserving semantically coherent anatomical structures. When the ODE constraints are removed, the performance degradation becomes more pronounced, with the PSNR decreasing to 27.11 dB on SPIDER and 34.23 dB on TopCow24, suggesting the importance of explicitly modeling inter-slice anatomical dynamics. As visualized in Fig.~\ref{feature evolution}, the instantaneous velocity fields enrich the average embeddings with shape-aware information, while semantic reweighting further enhances the reconstructed embeddings with intra-slice semantic correlations. In fact, this work's contribution is not the INR backbone itself, but the integration of inter-slice anatomical continuity and intra-slice semantic consistency into a spacing harmonization framework.

\noindent \textbf{\textit{Interaction between image resampling and image segmentation.}}
We further analyze the interaction between volumetric image resampling and downstream medical image segmentation. As discussed in Section~\ref{4.3}, unifying voxel spacing can reduce the distribution discrepancy caused by heterogeneous physical resolutions and thus improve segmentation robustness. Nevertheless, the effectiveness of spacing unification highly depends on the resampling quality. Conventional linear interpolation relies on intensity-level estimation, which may introduce anatomical discontinuities or blurred boundaries, especially for thin structures and complex anatomical regions.

In contrast, Consispace couples resampling with anatomical and semantic priors. The ODE-based interpolator models continuous inter-slice transitions through bidirectional instantaneous velocity fields, thereby preserving anatomical smoothness along the axial direction. Meanwhile, the DINOv3-guided intra-slice constraint introduces dense semantic correlations into the interpolation process. As shown in Fig.~\ref{feature evolution}, the semantic correlation maps tend to assign higher responses to voxels belonging to the same anatomical or segmentation class. By reweighting interpolated features with these correlations, Consispace enhances class-wise consistency and suppresses less relevant neighboring information. Therefore, Consispace does not treat resampling as an isolated preprocessing step, but establishes a beneficial interaction between resampling and segmentation: semantic priors improve resampling fidelity, while higher-quality resampled images further benefit downstream segmentation.

\begin{figure}[!t]
\centerline{\includegraphics[width=1.0\linewidth]{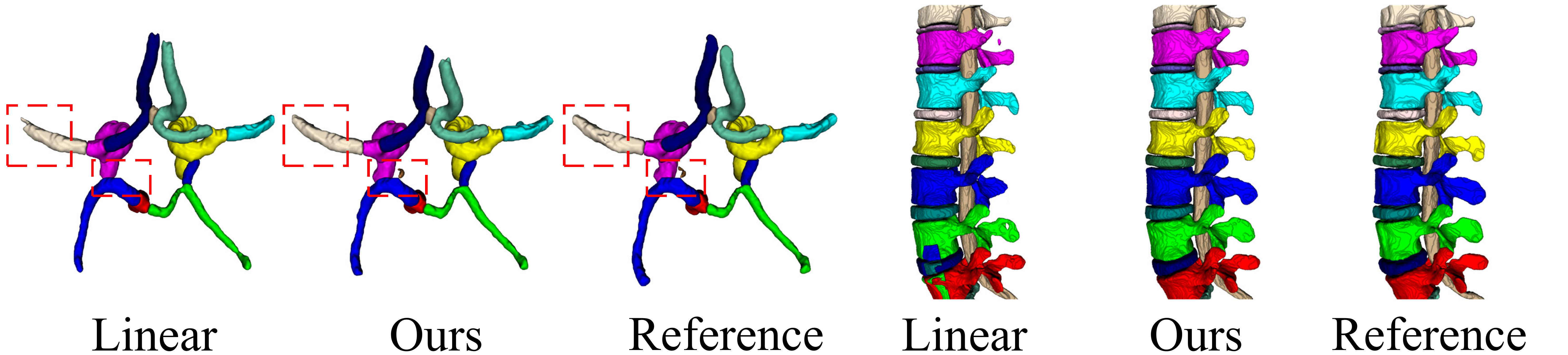}}
  \caption{The segmentation performance promotion when implementing Consispace on the finetuned MedSAM-2 segmentation model.
  }
  \label{sam_segmentation}
  \vspace{-4mm}
\end{figure}

\noindent \textbf{\textit{Application to SAM-based foundation models.}}
To examine the generality of Consispace beyond conventional segmentation architectures, we further apply it to the SAM-based medical foundation model MedSAM-2~\cite{zhu2024medical}. Specifically, we compare linear resampling and Consispace as preprocessing strategies before fine-tuning MedSAM-2. Quantitative results show that Consispace consistently improves the Dice score, from $84.62\%$ to $86.13\%$ on SPIDER and from $79.33\%$ to $81.89\%$ on TopCow24. These results indicate that the proposed resampling strategy is also beneficial for foundation models adapted to medical image segmentation.

The qualitative comparisons in Fig.~\ref{sam_segmentation} further demonstrate that Consispace produces segmentation results with more complete anatomical structures and more accurate boundary delineation. By preserving inter-slice anatomical continuity and intra-slice semantic consistency, Consispace provides more reliable volumetric inputs for MedSAM-2. These findings suggest that Consispace can serve as a general and model-agnostic preprocessing framework for medical image segmentation.

\section{Discussion}
\noindent$\bullet$ \textbf{\textit{Backbone selection for feature reweighting.}}
DINOv3 \cite{simeoni2025dinov3} is adopted in Consispace to construct intra-slice contextual correlations for reweighting the interpolated feature. To evaluate the influence of the feature backbone, we replace DINOv3 with several representative alternatives, including general-purpose pretrained vision models, i.e., MAE \cite{he2022masked} and DINOv2 \cite{oquab2023dinov2}, and medical foundation models, i.e., SAM-Med2D \cite{cheng2023sam} and VISTA3D \cite{he2025vista3d}. As reported in Table \ref{more degradation}, DINOv3 achieves the best overall resampling performance on SPIDER, with the highest PSNR of 28.31 dB and SSIM of 0.849. These results suggest that DINOv3 provides more reliable dense representations for identifying semantically correlated neighborhoods within each slice, thereby improving reconstruction fidelity. Compared with MAE and DINOv2, the medical foundation models generally yield better segmentation-related metrics, indicating the benefit of medical-domain priors.
\begin{table*}[!t]
  \begin{center}
  \caption{Ablation study on the hyperparameters of the kernel size and value k within the pretrained-vision-model guidance (The best results are highlighted in bold, and the second-best results are underlined).}
  \label{tab hyper-rameter}
  \resizebox{0.90\linewidth}{!}{
  \begin{tabular}{cccccccccc}  
\toprule 
 \multirow{2}*{Kernel size} &  \multirow{2}*{Value $k$} & \multicolumn{4}{c}{SPIDER} & \multicolumn{4}{c}{TopCow24}  \\  
  \cmidrule(r){3-6}  \cmidrule(r){7-10}  
&  & PSNR (dB) $\uparrow$ & SSIM $\uparrow$ & NMSE $\downarrow$ & LPIPS $\downarrow$ & PSNR (dB) $\uparrow$ & SSIM $\uparrow$ & NMSE $\downarrow$ & LPIPS $\downarrow$ \\
\midrule
$3 \times 3$  &  3 & 27.75 & 0.821 & 0.052 & 1.610 & 34.34 & 0.903 & 0.027 & 0.886 \\
$3 \times 3$  &  5  & 27.97 & 0.830 & 0.045 & 1.573 & 34.71  & 0.907 & 0.024 & 0.859   \\
$5 \times 5$  &  3  & 28.11 & 0.836 & \underline{0.040} & 1.469 & 34.82 & 0.909 & 0.020 &  0.875  \\
$5 \times 5$  &  5   & \underline{28.31} & \underline{0.849} & \textbf{0.039} & \textbf{1.418} & \underline{35.09} & \underline{0.916} & \underline{0.017} & \underline{0.768} \\
$5 \times 5$  &  7    & \textbf{28.40} & \textbf{0.851} & \underline{0.040} & \underline{1.432} & \textbf{35.17} & \textbf{0.919} & \textbf{0.016} &  \textbf{0.752} \\
$5 \times 5$  &  9    & 28.19 & 0.841 & 0.043 & 1.502 & 34.98 & 0.908 & 0.020 &  0.806 \\
\bottomrule 
  \end{tabular}}
  \end{center}
\end{table*}

\begin{table}[!t]
  \begin{center}
    \caption{Ablation study on the selection of backbones aimed at acquiring the contextual map. Quantitative results are implemented on the SPIDER dataset.}
\label{more degradation}
    \centering
    \resizebox{1.0\columnwidth}{!}{
    \begin{tabular}{lcccccc}
\toprule 
      \multirow{2}*{Model} & \multicolumn{4}{c}{Resampling} & \multicolumn{2}{c}{Segmentation} \\
      \cmidrule(r){2-5}    \cmidrule(r){6-7}  
    & PSNR $\uparrow$ & SSIM $\uparrow$ & NMSE $\downarrow$ & LPIPS $\downarrow$ & Dice $\uparrow$ & HD95 $\downarrow$  \\
\midrule
      DINOv2 \cite{oquab2023dinov2} & 27.63 & 0.814 & 0.052 & 1.535 & 88.87 & 5.38 \\
        MAE \cite{he2022masked} & 26.51 & 0.758 & 0.072 & 1.795 & 87.30 & 6.89 \\ 
                SAM-Med2D \cite{cheng2023sam} & 27.96 & 0.830 & 0.047 & 1.618 & 89.53 & 4.72 \\
        VISTA3D \cite{he2025vista3d} & \underline{28.04} & \underline{0.833} & \underline{0.041} & \textbf{1.359} & \textbf{90.97} & \underline{3.75} \\ 
        DINOv3 \cite{simeoni2025dinov3} & \textbf{28.31} & \textbf{0.849} & \textbf{0.039} & \underline{1.418} & \underline{90.56} & \textbf{3.48} \\
\bottomrule  
    \end{tabular}}
  \end{center}
  \vspace{-2mm}
\end{table}

Notably, adopting VISTA3D as the feature selector achieves the highest Dice score of $90.97\%$, which may be attributed to its strong 3D medical representations and robustness to point-prompt segmentation. However, since VISTA3D operates on volumetric inputs, it requires substantially higher GPU memory during feature extraction. In contrast, DINOv3 achieves the best reconstruction quality and the lowest HD95 of 3.48 mm while maintaining better computational efficiency. Therefore, we adopt DINOv3 as the default backbone for constructing intra-slice contextual maps.

\noindent$\bullet$ \textbf{\textit{Structural analysis of the ODE-based interpolator.}}
We further conduct ablation studies on the internal design of the ODE-based interpolator, including the directionality of ODE evolution and the attention formulation for instantaneous velocity fields. As reported in Table~\ref{tab ablation}, replacing the bidirectional ODE with a unidirectional variant consistently degrades reconstruction performance on both datasets. Specifically, the PSNR decreases from 28.31 dB to 27.97 dB on SPIDER and from 35.09 dB to 34.28 dB on TopCow24, with similar trends observed in SSIM, NMSE, and LPIPS. This indicates that single-directional evolution is insufficient to fully characterize axial anatomical variations, whereas bidirectional ODE modeling can integrate contextual information from both neighboring directions and better capture inter-slice transitions.

We also evaluate the attention formulation used for velocity field modeling. Replacing the proposed reversed attention with vanilla attention leads to clear performance drops, with the PSNR decreasing from 28.31 dB to 27.44 dB on SPIDER and from 35.09 dB to 34.56 dB on TopCow24. This suggests that vanilla attention tends to emphasize highly similar or relatively static regions, which is less suitable for modeling velocity fields that aim to capture inter-slice variations. In contrast, reversed attention highlights feature discrepancies between adjacent slices, enabling more accurate estimation of anatomical changes. These results validate the necessity of both bidirectional ODE evolution and reversed attention in the proposed interpolator.

\noindent$\bullet$ \textbf{\textit{Hyper-parameter analysis of the kernel size and value $k$.}} In this work, the local window is defined as a $5 \times 5$ square kernel centered at point $p$, and the value of $k$ in the proposed Consispace is set to $5$. We conduct ablation studies on both the kernel size and the value of $k$, as reported in Table \ref{tab hyper-rameter}. It can be observed that enlarging the local window from $3 \times 3$ to $5 \times 5$ consistently improves the reconstruction performance on both datasets. For example, when $k=5$, increasing the kernel size from $3 \times 3$ to $5 \times 5$ improves the PSNR from 27.97 dB to 28.31 dB on SPIDER, while also yielding better SSIM, NMSE, and LPIPS values. This indicates that a $5 \times 5$ local window can capture richer contextual information, thereby providing more reliable local consistency constraints.

Regarding the Top-$k$ operation, increasing $k$ from $3$ to $5$ brings clear performance gains under the $5 \times 5$ kernel setting. Although setting $k$ to $7$ achieves the best PSNR and SSIM on both datasets, the improvement over $k=5$ is marginal. Meanwhile, a larger $k$ inevitably introduces additional computational cost due to the increased number of selected candidates. Moreover, further increasing $k$ to $9$ leads to degraded performance, suggesting that incorporating excessive candidates will introduce more irrelevant information. Therefore, we set the kernel size to $5 \times 5$ and $k$ to $5$, which achieves a favorable trade-off between reconstruction accuracy and computational efficiency.

\section{Conclusion and Future Work}
We systematically study the interplay between medical image resampling and segmentation, and propose Consispace, a unified resampling framework that couples upstream reconstruction with downstream segmentation. Consispace incorporates ODE-based anatomical constraints to model inter-slice dynamics, alleviating the limitations of conventional discrete resampling. Moreover, we leverage a pretrained vision model to construct intra-slice correlation maps, injecting class-wise semantic consistency through feature reweighting during resampling. Both intra-slice and inter-slice constraints are integrated into an INR formulation. Extensive experiments demonstrate that Consispace improves reconstruction quality and perceptual fidelity, produces smoother inter-slice anatomical variations, and boosts downstream segmentation performance when used as a preprocessing step. However, this study adopts an anatomy-specific setting rather than a universal resampling-segmentation model. For future work, we will scale up both model capacity and training data diversity to improve generalization to unseen anatomical domains.

\bibliographystyle{IEEEtran}
\bibliography{ref}


\end{document}